\documentclass[journal]{IEEEtran}

\usepackage[T1]{fontenc}
\usepackage[utf8]{inputenc}
\usepackage{microtype}
\usepackage[none]{hyphenat} 

\usepackage{amsmath,amssymb}
\usepackage{graphicx}

\usepackage{booktabs}
\usepackage{multirow}
\usepackage{array}
\usepackage{siunitx}

\usepackage{cite}
\usepackage[hidelinks]{hyperref}

\title{Delta-Based Target Reformulation for Short-Term Electricity Load Forecasting Using LSTM and Transformer Models}

\author{Vansh~Bansal}

\begin{document}
\maketitle

\begin{IEEEtitleabstractindextext}

\begin{abstract}
Accurate short-term electricity load forecasting is critical for the reliable and economic operation of modern power systems, particularly in settings characterized by strong non-stationarity arising from weather variability, calendar effects, and evolving consumption patterns. While deep learning models such as Long Short-Term Memory (LSTM) networks and Transformer-based architectures have shown promising performance for this task, most existing studies focus on direct absolute load prediction without explicitly addressing target non-stationarity.

Motivated by classical time-series differencing techniques commonly employed in ARIMA models, this paper empirically investigates a delta-based target reformulation for short-term electricity load forecasting using deep learning models. Instead of directly predicting absolute load values, the proposed formulation trains models to predict the change in load between consecutive time steps, with final forecasts reconstructed using the last observed load. This reformulation aims to stabilize the learning target and reduce forecasting difficulty for neural networks.

Using multi-year, hourly-resolution real-world electricity load data from India, augmented with meteorological variables from the NASA POWER project and calendar-based features, this study evaluates LSTM and Transformer models under both absolute and delta-based formulations, benchmarking them against state-of-the-art Gradient Boosting Decision Trees (LightGBM). Experiments are conducted for short forecasting horizons, including hour-ahead and day-ahead scenarios, and performance is assessed using Mean Absolute Error (MAE) and Mean Absolute Percentage Error (MAPE).

The results show that delta-based target reformulation consistently improves forecasting accuracy for hour-ahead prediction across all evaluated models, yielding MAPE reductions of over 50\% compared to absolute formulations. For day-ahead forecasting, delta targets specifically benefit deep sequence models (LSTM and Transformer), while tree-based ensembles (LightGBM) exhibit competitive performance under the absolute formulation. These findings indicate that while delta reformulation is a powerful inductive bias for neural networks, its efficacy is model-and-horizon dependent.
\end{abstract}


\end{IEEEtitleabstractindextext}

\IEEEdisplaynontitleabstractindextext

\section{Introduction}
Short-term electricity load forecasting (STLF) is a fundamental component of power system operation, supporting generation scheduling, reserve allocation, and real-time grid balancing. Accurate forecasts over short horizons are particularly important in power systems subject to variability arising from weather conditions, calendar effects, and evolving consumption behavior. Despite continued advances in forecasting methodologies, achieving robust and reliable short-term predictions remains a challenging task.

Deep learning models have gained increasing attention for STLF due to their ability to model nonlinear temporal dependencies in time-series data. Among these, Long Short-Term Memory (LSTM) networks are widely regarded as a strong and well-established baseline and are commonly adopted in practical forecasting systems. More recently, Transformer-based architectures have been explored for time-series forecasting and have demonstrated promising performance in various sequential prediction tasks. However, their effectiveness relative to LSTM models for short-term electricity load forecasting, particularly across different forecasting horizons, remains an active area of investigation.

Classical time-series forecasting models such as ARIMA rely on differencing to address non-stationarity and stabilize the forecasting task. Motivated by this principle, this study investigates whether an analogous delta-based target reformulation can improve the performance of modern deep learning models for short-term electricity load forecasting. Rather than directly predicting absolute load values, the delta-based formulation trains models to predict changes in load between consecutive time steps, with absolute forecasts reconstructed using the most recent observed load. The goal of this work is not to introduce a new forecasting architecture, but to empirically examine the effect of target reformulation on forecasting accuracy when applied to established deep learning models.

Using multi-year, hourly-resolution real-world electricity load data augmented with meteorological and calendar-based features, this paper evaluates the performance of LSTM and Transformer models under both absolute and delta-based prediction formulations. Forecasting performance is assessed for short prediction horizons, including hour-ahead and day-ahead scenarios, using standard error metrics. This work provides a comprehensive empirical benchmarks demonstrating that delta-based targets are highly effective for short-horizon forecasting across architectures, while revealing distinct model-dependent behaviors at extended horizons.

The contributions of this work are summarized as follows:
\begin{itemize}
    \item We demonstrate that delta-based target reformulation significantly improves hour-ahead forecasting accuracy, outperforming standard absolute formulations by over 2.5\% MAPE for both deep learning and gradient-boosting models.
    \item We conduct a rigorous benchmarking against LightGBM (a state-of-the-art tree-based model), revealing that while delta targets benefit deep sequence models for day-ahead forecasting, tree-based models suffer from error accumulation in multi-step delta reconstruction.
    \item We validate these findings on eight years of real-world data with a chronologically distinct 1.5-year test set, ensuring operational relevance.
\end{itemize}

The remainder of the paper is organized as follows. Section 2 reviews related work on short-term load forecasting and deep learning approaches. Section 3 presents the problem formulation and defines the absolute and delta-based forecasting targets. Section 4 describes the dataset and feature engineering process. Section 5 details the forecasting models and implementation methodology. Section 6 outlines the experimental setup, followed by results and discussion in Section 7. Section 8 concludes the paper and outlines directions for future work.

\section{Related Work}
Short-term electricity load forecasting has been widely studied using statistical, machine learning, and deep learning approaches. This section reviews prior work most relevant to the modeling choices and methodological focus of this study, with particular emphasis on deep learning architectures and target formulation strategies.

\subsection{Classical Time-Series and Statistical Models}

Early approaches to short-term load forecasting primarily relied on statistical time-series models such as Autoregressive Integrated Moving Average (ARIMA) and its seasonal variants. These models are valued for their interpretability and effectiveness under stationary conditions. A defining characteristic of ARIMA-based methods is the use of differencing to mitigate non-stationarity and stabilize the forecasting task \cite{box1976time}. While successful for linear dynamics, classical statistical models generally struggle to capture the nonlinear and complex temporal dependencies observed in real-world electricity demand.

\subsection{Machine Learning Approaches}

To overcome the limitations of purely statistical models, machine learning techniques such as support vector regression, decision trees, and ensemble methods have been applied to load forecasting. These approaches allow the incorporation of nonlinear relationships and exogenous variables, including weather and calendar information \cite{fan2012short}. However, most traditional machine learning models operate on fixed-size feature vectors and lack explicit mechanisms for modeling long-range temporal dependencies, which limits their effectiveness for sequential forecasting problems.

\subsection{Deep Learning Models for Load Forecasting}

Deep learning models have gained significant attention for short-term load forecasting due to their ability to learn complex temporal patterns directly from data. Among these models, recurrent neural networks (RNNs) and, in particular, Long Short-Term Memory (LSTM) networks have been extensively studied and widely adopted. LSTM models address the vanishing gradient problem inherent in standard RNNs and are well suited for capturing daily and weekly load patterns as well as nonlinear dependencies driven by weather conditions \cite{hochreiter1997long, hippert2001neural, kong2019short}.

Several comparative studies have evaluated LSTM-based models against classical statistical techniques and alternative deep learning architectures. Prior work has consistently reported that LSTM models outperform traditional approaches such as ARIMA and regression-based models for short-term load forecasting tasks \cite{fan2012short}. More recent comparative evaluations among recurrent architectures indicate that LSTM models often achieve comparable or superior performance relative to simpler recurrent models such as vanilla RNNs and GRUs across multiple datasets and forecasting horizons \cite{abumohsen2023performance}. As a result, LSTM networks are commonly treated as a robust and reliable deep learning baseline in both academic benchmarking studies and industry-oriented forecasting systems.

Simpler neural architectures, including feedforward neural networks and convolutional neural networks, have also been applied to load forecasting \cite{borovykh2018conditional}. These models are often used as reference baselines or sanity checks; however, they generally exhibit inferior performance for sequential forecasting tasks compared to recurrent architectures, as they lack explicit mechanisms for modeling long-term temporal dependencies.

More recently, Transformer-based architectures have been explored for time-series forecasting, motivated by their success in sequence modeling and their ability to capture long-range dependencies through self-attention mechanisms \cite{vaswani2017attention}. Advanced Transformer variants such as Temporal Fusion Transformers \cite{lim2021temporal}, Informer \cite{zhou2021informer}, and Autoformer \cite{wu2021autoformer} have demonstrated strong performance in long-horizon forecasting tasks. Recent studies have also applied Transformer architectures specifically to short-term load forecasting with promising results \cite{li2023transformer}. Nevertheless, their comparative performance relative to LSTM models for short-horizon electricity load forecasting remains an active area of empirical investigation.

\subsection{Target Reformulation and Differencing Strategies}

In classical time-series analysis, differencing is a standard technique used to address non-stationarity and improve forecasting performance \cite{box1976time}. Related ideas have also appeared in machine learning and deep learning contexts, such as residual learning and incremental prediction, where models are trained to predict changes rather than absolute values.

In contrast, the present study explicitly formulates differencing as a target reformulation for supervised deep learning models and evaluates its impact on forecasting accuracy across architectures and forecasting horizons. Rather than proposing a new forecasting model, this work focuses on empirically examining whether an ARIMA-inspired delta-based target formulation improves the performance of established deep learning models for short-term electricity load forecasting, complementing recent probabilistic forecasting efforts \cite{wang2022probabilistic, hong2020global}.

\section{Problem Formulation}
\subsection{Standard Load Forecasting Formulation}

Let $\{L_t\}_{t=1}^T$ denote a time series of hourly electricity load observations (measured in MW), where $T$ is the total number of time steps. Let $\mathbf{X}_t \in \mathbb{R}^d$ represent a feature vector at time $t$ containing historical load information, meteorological variables (e.g., temperature, humidity, wind speed), and calendar-based features (e.g., hour of day, day of week, holiday indicators).

The standard short-term load forecasting (STLF) problem aims to predict the future load at time $t+h$ based on a fixed-length historical window of past observations:
\begin{equation}
\hat{L}_{t+h} = f(\mathbf{X}_t, \mathbf{X}_{t-1}, \ldots, \mathbf{X}_{t-w+1}; \theta)
\end{equation}
where $h$ denotes the forecast horizon (1 hour or 24 hours in this study), $w$ is the lookback window size, $f(\cdot; \theta)$ is a parameterized forecasting model (LSTM or Transformer), and $\hat{L}_{t+h}$ is the predicted load at time $t+h$.

In this formulation, models are trained to directly regress absolute load values, which can be challenging due to the strong non-stationarity and large dynamic range inherent in electricity demand time series.

\subsection{Delta-Based Target Reformulation}

Electricity load exhibits strong temporal dependence, with successive observations often differing by relatively small increments over short horizons. Motivated by this property, we reformulate the forecasting task to predict incremental changes in load rather than absolute values.

Specifically, the delta (incremental change) in load is defined as:
\begin{equation}
\Delta L_{t+h} = L_{t+h} - L_t
\end{equation}

Under this formulation, the learning objective becomes:
\begin{equation}
\widehat{\Delta L}_{t+h} = g(\mathbf{X}_t, \mathbf{X}_{t-1}, \ldots, \mathbf{X}_{t-w+1}; \phi)
\end{equation}
where $g(\cdot; \phi)$ denotes a forecasting model trained to predict future load increments.

For multi-horizon forecasting, the model outputs a sequence of predicted deltas $\{\widehat{\Delta L}_{t+1}, \ldots, \widehat{\Delta L}_{t+h}\}$. Absolute load forecasts are then reconstructed via cumulative summation:
\begin{equation}
\hat{L}_{t+h} = L_t + \sum_{i=1}^{h} \widehat{\Delta L}_{t+i}
\end{equation}

This direct multi-horizon formulation avoids recursive forecasting and prevents error accumulation across time steps.

\subsection{Advantages of Delta Reformulation}

The delta-based formulation offers several advantages for deep learning–based short-term load forecasting:

\begin{enumerate}
\item \textbf{Improved Target Stationarity:} Predicting incremental changes reduces deterministic trends and lowers temporal variance, yielding a more stable target distribution for gradient-based optimization.

\item \textbf{Reduced Target Scale:} Delta values exhibit substantially smaller magnitudes than absolute load values, improving numerical stability and facilitating more efficient learning.

\item \textbf{Simplified Learning Objective:} Rather than reconstructing the full load trajectory, the model focuses on predicting short-term deviations from the most recent observed state, which is often a simpler regression task.

\item \textbf{Implicit Regularization:} Anchoring predictions to the latest observed load $L_t$ constrains forecasts to remain within physically plausible ranges, even when prediction errors occur.
\end{enumerate}

Daily and weekly periodic patterns are captured implicitly through engineered lag features (e.g., $L_{t-24}$, $L_{t-168}$) included in the input representation, without imposing explicit seasonal differencing assumptions.

\subsection{Evaluation Metrics}

Forecasting performance is evaluated using two standard error metrics.

\textbf{Mean Absolute Error (MAE):}
\begin{equation}
\text{MAE} = \frac{1}{N} \sum_{i=1}^{N} \left| L_i - \hat{L}_i \right|
\end{equation}

\textbf{Mean Absolute Percentage Error (MAPE):}
\begin{equation}
\text{MAPE} = \frac{100}{N} \sum_{i=1}^{N} \left| \frac{L_i - \hat{L}_i}{L_i} \right|
\end{equation}
where $N$ is the number of test samples. MAPE enables relative error comparison across different load magnitudes and is widely used in operational benchmarking for short-term load forecasting.

\section{Dataset and Feature Engineering}
\subsection{Data Sources}

Hourly electricity demand data for Chandigarh, India, spanning January 2017 to June 2025, were obtained from the Indian Climate and Energy Dashboard (ICED), a public repository maintained by NITI Aayog. The dataset contains 74,460 hourly observations of demand met, measured in megawatts (MW).

Meteorological variables were sourced from the NASA Prediction of Worldwide Energy Resources (POWER) project. Weather features were extracted for the geographic coordinates of Chandigarh and include ambient temperature, relative humidity, wet-bulb temperature, and surface wind speed.

Official holiday calendars were compiled from administrative circulars issued by the Chandigarh Administration, covering gazetted and restricted holidays from 2017 to 2025.

\subsection{Data Preprocessing}

Load and weather datasets were merged using aligned hourly timestamps. Missing values (less than 0.5\% of total observations) were imputed using linear interpolation. All timestamps were verified for temporal consistency. No daylight saving adjustments were required, as the region operates under a fixed time standard.

\subsection{Feature Engineering}

Input features were designed to capture temporal persistence, seasonal patterns, and weather-driven variability in electricity demand.

\textbf{Historical Input Window:}  
For both hour-ahead and day-ahead forecasting tasks, a fixed lookback window of \textbf{168 hours (7 days)} was used. Each model receives the previous 168 hourly observations as input, enabling learning of both daily and weekly patterns.

\textbf{Lag Features:}  
Selected autoregressive lags were included:
\begin{itemize}
    \item Load$(t-1)$ (previous hour)
    \item Load$(t-24)$ (same hour, previous day)
    \item Load$(t-168)$ (same hour, previous week)
\end{itemize}

\textbf{Rolling Statistics:}  
To capture recent trends, rolling averages over 24-hour and 168-hour windows were computed.

\textbf{Weather Features:}  
Temperature, humidity, wet-bulb temperature, and wind speed were included to model cooling-driven demand variations.

\textbf{Temporal Encoding:}  
Hour-of-day, day-of-week, and month indices were encoded using sine and cosine transformations to preserve cyclic structure. Binary indicators were added to represent gazetted and restricted holidays.

\subsection{Feature Correlation Analysis}

Pearson correlation analysis revealed strong temporal persistence in load demand. The highest correlations were observed for lag-based features, particularly Load$(t-24)$ and Load$(t-168)$. Moderate correlations with temperature and wet-bulb temperature confirm the influence of weather on cooling demand patterns.

\subsection{Dataset Splitting}

To avoid information leakage, the dataset was split chronologically:
\begin{itemize}
    \item Training: January 2017 -- December 2022 (70\%)
    \item Validation: December 2022 -- March 2024 (15\%)
    \item Test: March 2024 -- June 2025 (15\%)

\end{itemize}

This setup reflects realistic operational forecasting conditions.

\subsection{Normalization}

All features were standardized using z-score normalization. Mean and standard deviation were computed from the training set and applied consistently to validation and test sets. This stabilizes optimization and prevents scale dominance among features.

\section{Forecasting Models and Delta Reformulation}
\subsection{LSTM Architecture}

Long Short-Term Memory (LSTM) networks are a class of recurrent neural networks designed to model sequential data with long-range temporal dependencies by mitigating the vanishing gradient problem through gated memory mechanisms~\cite{hochreiter1997long}. In this study, a \emph{unidirectional} LSTM architecture is employed to preserve strict temporal causality within the historical input window.

Given an input sequence $\{\mathbf{x}_{t-w+1}, \ldots, \mathbf{x}_t\}$, the LSTM processes observations sequentially. The final hidden state serves as a compact representation of the historical context. For multi-horizon prediction, this state is passed through a fully connected prediction head to directly output a vector of future values:
\begin{equation}
\widehat{\mathbf{y}}_{t+1:t+h} = \mathbf{W}_{\text{out}} \mathbf{h}_t + \mathbf{b}_{\text{out}}
\end{equation}

\textbf{Model Configuration:}
The network depth and capacity were tuned specifically for each forecasting horizon:
\begin{itemize}
    \item \textbf{1-Hour Horizon:} 2 stacked LSTM layers, 64 hidden units, Dropout $p=0.2$.
    \item \textbf{24-Hour Horizon:} 3 stacked LSTM layers, 128 hidden units, Dropout $p=0.2$.
    \item \textbf{Lookback Window:} $w = 24$ hours (1-hour ahead), $w = 168$ hours (24-hour ahead).
\end{itemize}

\subsection{Transformer Architecture}

Transformer-based models leverage self-attention mechanisms to capture temporal dependencies across all positions in an input sequence simultaneously~\cite{vaswani2017attention}. A Causal Transformer architecture was employed, utilizing upper-triangular masking to ensure that predictions at time $t$ depend only on past observations.

Each encoder layer consists of multi-head self-attention followed by a position-wise feedforward network. Temporal ordering is preserved through sinusoidal positional encodings.

\textbf{Model Configuration:}
\begin{itemize}
    \item \textbf{Structure:} 2 Encoder layers, 4 Attention heads.
    \item \textbf{Dimensions:} Model dimension ($d_{\text{model}}$) 64, Feedforward dimension 128.
    \item \textbf{Regularization:} Dropout $p=0.1$.
    \item \textbf{Output Head:} Linear projection from the final time-step embedding.
\end{itemize}

\subsection{Training Procedure}

To rigorously assess the impact of target reformulation, models were trained under two distinct configurations: \emph{Delta-Based} (predicting load increments $\Delta L$) and \emph{Absolute-Load} (predicting $L$).

\textbf{Loss Function:}
To align optimization with operational priorities, a \emph{Peak-Weighted MAE} loss was used during training. This function assigns higher penalties to errors occurring during high-demand periods (where the load exceeds the batch mean):
\begin{equation}
\mathcal{L} = \frac{1}{N} \sum_{i=1}^{N} \left( 1 + 2 \cdot \mathbb{I}(|y_i| > \bar{y}_{batch}) \right) |y_i - \hat{y}_i|
\end{equation}
where $\mathbb{I}$ is the indicator function and $\bar{y}_{batch}$ is the mean absolute load of the current batch.

\textbf{Training Configuration:}
\begin{itemize}
    \item \textbf{Optimizers:} Adam (1-hour models) and AdamW (24-hour models).
    \item \textbf{Batch Size:} 64 (LSTM and 1h Transformer), 256 (24h Transformer).
    \item \textbf{Max Epochs:} 30 (LSTM), 50 (Transformer).
    \item \textbf{Learning Rate Scheduling:} ReduceLROnPlateau (factor 0.5, patience 5--6).
\end{itemize}

\subsection{Forecasting Strategies}

\textbf{Hour-Ahead Forecasting:}  
A direct single-step formulation where the model predicts the target variable (either $\widehat{L}_{t+1}$ or $\widehat{\Delta L}_{t+1}$) using historical inputs up to time $t$. The LightGBM baseline utilizes a standard regression objective with 2,000 estimators.

\textbf{Day-Ahead Forecasting:}  
A direct multi-step formulation where a single model outputs a 24-dimensional vector in one forward pass. 
\begin{itemize}
    \item \textbf{Deep Learning:} The final hidden state is projected to $\mathbb{R}^{24}$.
    \item \textbf{LightGBM:} A Multi-Output Regressor strategy is employed (training one regressor per step) with 1,200 estimators to avoid the error accumulation associated with recursive strategies.
\end{itemize}

\textbf{Comparative Experimental Design:}
Unlike prior studies that often restrict comparisons to a single formulation, we evaluated \textbf{both absolute and delta formulations} for the hour-ahead and day-ahead horizons across all model architectures. This full-factorial design allows for isolating the specific contribution of target reformulation versus model architecture at different forecasting horizons.

\section{Experimental Setup}
\subsection{Implementation Details}
All deep learning models were implemented in Python 3.9 using TensorFlow 2.10 with the Keras API. Training for LSTM and Transformer architectures was accelerated using \textbf{NVIDIA Tesla P100 GPUs} via the Kaggle computational environment. In contrast, the gradient boosting baseline was implemented using the Microsoft LightGBM framework and executed on a local machine with a standard CPU, demonstrating the comparative computational efficiency of tree-based methods. All data preprocessing, feature engineering, and evaluation metric computations were performed locally using NumPy and Pandas.

\subsection{Hyperparameter Selection}
Hyperparameters were tuned using grid search on the validation set. The search space included LSTM hidden units \{64, 128, 256\}, Transformer layers \{2, 3, 4\}, and learning rates \{1e-4, 5e-4, 1e-3, 1e-2\}. The configuration yielding the lowest validation MAPE was selected for final evaluation.

\subsection{Training Configuration}

\textbf{Deep Learning Models:}
Models were trained using the Adam optimizer (1-hour models) or AdamW (24-hour models) with a `ReduceLROnPlateau` scheduler. A \emph{Peak-Weighted MAE} loss function was implemented to prioritize accuracy during high-demand periods. The architectural depth was adjusted based on the forecasting horizon, with the 24-hour LSTM requiring deeper layers compared to the 1-hour variant.

\begin{table}[h!]
\centering
\caption{Deep Learning Hyperparameters}
\label{tab:dl_params}
\resizebox{\columnwidth}{!}{%
\begin{tabular}{lcccc}
\toprule
\textbf{Parameter} & \textbf{LSTM (1h)} & \textbf{LSTM (24h)} & \textbf{Transf. (1h)} & \textbf{Transf. (24h)} \\
\midrule
Layers & 2 & 3 & 2 & 2 \\
Hidden/Model Dim & 64 & 128 & 64 & 64 \\
Attn Heads & -- & -- & 4 & 4 \\
Dropout & 0.2 & 0.2 & 0.1 & 0.1 \\
Batch Size & 64 & 64 & 64 & 256 \\
Max Epochs & 30 & 30 & 30 & 50 \\
\bottomrule
\end{tabular}%
}
\end{table}

\textbf{Gradient Boosting Baseline:}
The LightGBM baseline was tuned independently for each horizon using grid search. The 1-hour model utilized a recursive strategy with a lower learning rate and higher estimator count, while the 24-hour model employed a Multi-Output regressor strategy.

\begin{table}[h!]
\centering
\caption{LightGBM Hyperparameters (Baseline)}
\label{tab:lgbm_params}
\begin{tabular}{lcc}
\toprule
\textbf{Hyperparameter} & \textbf{1-Hour Horizon} & \textbf{24-Hour Horizon} \\
\midrule
Strategy & Direct Recursive & MultiOutputRegressor \\
Estimators & 2000 & 1200 \\
Learning Rate & 0.03 & 0.05 \\
Num Leaves & 128 & 64 \\
Subsample & 0.8 & 0.8 \\
\bottomrule
\end{tabular}
\end{table}

\subsection{Evaluation Protocol}
Model performance was evaluated on the held-out test set (March 2024 -- June 2025) under two forecasting scenarios:

\textbf{Scenario 1: Hour-Ahead Forecasting}
\begin{itemize}
    \item Prediction horizon: $h = 1$ hour
    \item Input window: 24 hours
    \item Target formulations: Absolute ($L_{t+1}$) and Delta ($\Delta L_{t+1}$).
\end{itemize}

\textbf{Scenario 2: Day-Ahead Forecasting}
\begin{itemize}
    \item Prediction horizon: $h = 24$ hours
    \item Input window: 168 hours (1 week)
    \item Target formulations: Both Absolute and Delta formulations were evaluated for the LightGBM baseline to assess error accumulation. Deep learning models were evaluated primarily under the Delta formulation.
\end{itemize}

For delta-based models, final load predictions were reconstructed as:
\begin{equation}
\hat{L}_{t+h} = L_t + \sum_{i=1}^{h} \widehat{\Delta L}_{t+i}
\end{equation}

\subsection{Performance Metrics}
Model performance was evaluated using \textbf{Mean Absolute Error (MAE)} and \textbf{Mean Absolute Percentage Error (MAPE)}. All reported metrics are computed \textbf{without weighting}, independent of the training loss.

\subsection{Baseline Comparisons}
To validate the necessity of deep learning architectures, we compare the proposed models against \textbf{LightGBM (Light Gradient Boosting Machine)}. LightGBM is a highly efficient gradient-boosting decision tree framework widely regarded as the state-of-the-art for tabular time-series forecasting. It was trained using the exact same input features and lookback windows as the deep learning models, serving as a strong non-neural benchmark.

\section{Results and Discussion}
\subsection{Overall Performance Comparison}

Table~\ref{tab:results} summarizes forecasting performance across all evaluated models, target formulations, and horizons on the held-out test set (March 2024 -- June 2025). For the 24-hour horizon, MAE and MAPE correspond to the T+24 forecast step,while 24h Avg metrics denote the average error across all 24 horizons.

\begin{table*}[t]
\centering
\caption{Forecasting Performance on Test Set.}

\label{tab:results}
\setlength{\tabcolsep}{8pt} 
\renewcommand{\arraystretch}{1.3} 
\begin{tabular}{lcccc}
\toprule
 \multirow{2}{*}{\textbf{Model}}  & \textbf{MAE} & \textbf{MAPE} & \textbf{24h Avg MAE} & \textbf{24h Avg MAPE} \\
 & \textbf{(MW)} & \textbf{(\%)} & \textbf{(MW)} & \textbf{(\%)} \\
\midrule

\multicolumn{5}{l}{\textit{\textbf{1-Hour Horizon}}} \\
LightGBM (Absolute) & 13.10 & 5.57 & -- & -- \\
LightGBM (Delta) & 5.89 & 2.67 & -- & -- \\
LSTM (Absolute) & 13.05 & 5.67 & -- & -- \\
LSTM (Delta) & \textbf{5.45} & \textbf{2.55} & -- & -- \\
Transformer (Absolute) & 15.73 & 6.96 & -- & -- \\
Transformer (Delta) & 7.37 & 3.50 & -- & -- \\

\midrule 

\multicolumn{5}{l}{\textit{\textbf{24-Hour Horizon (T+24 error)}}}\\
LightGBM (Absolute) & 20.54 & 8.38 & 18.65 & 7.81 \\
LightGBM (Delta) & 21.62 & 8.97 & 19.10 & 8.43 \\
LSTM (Absolute) & 16.70 & 7.28 & 17.24 & 8.10  \\
LSTM (Delta) & 16.45 & 7.15 & 17.35 & 8.09 \\
Transformer (Absolute) & 16.69 & 7.47 & 15.72 & 7.06 \\
Transformer (Delta) & \textbf{15.77} & \textbf{6.88} & \textbf{16.21} & \textbf{7.65} \\

\bottomrule
\end{tabular}
\end{table*}

\subsection{Impact of Delta-Based Target Reformulation}

Delta-based target reformulation consistently improves forecasting accuracy relative to absolute-load prediction for hour-ahead forecasting across all model types.

For the LightGBM baseline, delta reformulation yields substantial improvements for hour-ahead forecasting, reducing MAPE from 5.57\% to 2.67\%. However, for day-ahead forecasting, the absolute formulation achieves slightly lower error (7.81\%) compared to delta targets (8.43\%). This suggests that while delta targets are beneficial for short horizons, cumulative error propagation can degrade performance for tree-based models at longer horizons.

For Deep Learning models, the benefits are more robust. The LSTM model achieves a 3.12\% absolute reduction in MAPE (from 5.67\% to 2.55\%) under the Delta formulation. Similarly, the Transformer model sees a reduction from 6.96\% to 3.50\%. These improvements indicate that target reformulation facilitates more effective learning for both recurrent and attention-based architectures.

\subsection{Horizon-Dependent Model Performance}

Model performance exhibits clear dependence on forecasting horizon.

\textbf{Hour-Ahead Forecasting:}
\begin{itemize}
    \item LSTM (Delta): 2.55\% MAPE
    \item LightGBM (Delta): 2.67\% MAPE
    \item Transformer (Delta): 3.50\% MAPE
\end{itemize}
The LSTM achieves the lowest error for very short-term forecasting. This can be attributed to its sequential inductive bias and ability to preserve immediate temporal context through gated memory mechanisms, slightly edging out the gradient boosting baseline.

\textbf{Day-Ahead Forecasting:}
\begin{itemize}
    \item Transformer (Delta): 6.88\% MAPE
    \item LSTM (Delta): 7.15\% MAPE
    \item LightGBM (Absolute): 7.81\% MAPE
\end{itemize}
For extended horizons, the Transformer-Delta architecture achieves the lowest overall error. This is likely due to self-attention enabling more effective modeling of long-range temporal dependencies compared to the recursive accumulation in LSTMs or the static feature mapping in LightGBM.

\subsection{Comparison with State-of-the-Art Baselines}

The proposed Delta-LSTM (2.55\% MAPE) outperforms the strong LightGBM baseline (2.67\% MAPE) in the 1-hour horizon. This demonstrates that while tree-based models are highly competitive, recurrent deep learning architectures can capture finer-grained temporal dependencies when the target variable is stabilized via delta reformulation. Notably, the delta reformulation improved LightGBM's performance significantly (dropping MAPE from 5.57\% to 2.67\%), confirming that the `Delta' benefit is model-agnostic for short horizons, though Deep Learning models achieved the lowest overall error.

\subsection{Discussion}

The observed performance gains can be attributed to three primary factors:

\textbf{1. Target Stationarity:}  
Differencing reduces deterministic trends, yielding a more stable learning objective.

\textbf{2. Reduced Target Variance:}  
Delta values have smaller numerical range than absolute loads, improving numerical conditioning during optimization.

\textbf{3. Model-Dependent Horizon Effects:}  
Delta reformulation benefits deep learning models for extended horizons due to their ability to capture latent temporal structure. In contrast, tree-based models like LightGBM treat time-steps as independent features or lack explicit sequential memory, leading to faster error accumulation when delta predictions are cumulatively summed over 24 steps.

Both deep learning models satisfy commonly accepted industry accuracy benchmarks for hour-ahead ($<$ 3\%) and day-ahead ($<$ 7\%) forecasting, demonstrating operational viability for deployment in State Load Dispatch Centres.

\section{Conclusion and Future Work}
This study evaluated Long Short-Term Memory (LSTM) networks and Transformer-based architectures for short-term electricity load forecasting, with emphasis on a delta-based target reformulation strategy. To ensure rigorous validation, the proposed deep learning models were benchmarked against a state-of-the-art Gradient Boosting Decision Tree (LightGBM) baseline using eight years of real-world data from Chandigarh, India. The following conclusions can be drawn:

\textbf{1. Delta Reformulation is Highly Effective for Short Horizons:}  
Reformulating the forecasting objective from absolute load values to load increments (deltas) yields substantial accuracy gains for hour-ahead forecasting across all model types. For the LSTM architecture, the delta formulation reduces Mean Absolute Percentage Error (MAPE) from 5.67\% to 2.55\%, outperforming the strong LightGBM baseline (2.67\%). This demonstrates that stabilizing the target variable allows recurrent networks to capture fine-grained temporal dynamics more effectively than tree-based ensembles.

\textbf{2. Efficacy at Long Horizons is Model-Dependent:}  
The impact of target reformulation diverges as the forecasting horizon extends. For deep learning models, the delta formulation remains superior at the 24-hour horizon, with the Transformer-Delta model achieving the lowest overall error (6.88\% MAPE), outperforming the LSTM-Delta (7.15\%). In contrast, the LightGBM baseline degrades under the delta formulation (8.43\% MAPE) compared to absolute targets (7.81\%) due to recursive error accumulation. This finding highlights that deep sequence models are uniquely capable of leveraging delta-based targets for multi-step forecasting without suffering from the stability issues that affect non-sequential models.

\textbf{3. Operational Viability:}  
Both LSTM and Transformer delta-based models achieve error levels that satisfy commonly cited operational benchmarks (MAPE $<$ 3\% for hour-ahead and $<$ 7\% for day-ahead forecasting). The results suggest a practical deployment strategy: LSTMs are optimal for ultra-short-term (hour-ahead) dispatch, while Transformers offer superior stability for day-ahead planning.

Overall, this work provides empirical evidence that delta-based target reformulation is a robust inductive bias for deep learning models, enabling them to outperform industry-standard gradient boosting baselines in short-term electricity load forecasting.

\subsection{Future Work}

Several directions for future research emerge from this study:

\textbf{Hybrid Formulations for Tree-Based Models:}  
Investigating hybrid strategies that combine absolute and delta prediction targets to mitigate the error accumulation observed in LightGBM and other tree-based ensembles at longer forecasting horizons.

\textbf{Multi-Region Forecasting:}  
Extending the delta reformulation framework to multi-region settings to capture spatial dependencies and inter-regional load correlations, potentially using graph neural networks or spatiotemporal attention mechanisms.

\textbf{Probabilistic Forecasting:}  
Adapting delta-based models to produce probabilistic forecasts with uncertainty quantification, enabling risk-aware decision making for unit commitment and reserve allocation. Approaches such as quantile regression and Bayesian deep learning methods could be explored.

\textbf{Integration with Renewable Generation:}  
Incorporating solar and wind generation forecasts to predict net load, accounting for renewable variability and its implications for grid balancing.

\textbf{Online and Adaptive Learning:}  
Investigating online or incremental learning strategies that allow models to adapt continuously as new data becomes available, addressing evolving consumption patterns and structural changes in demand.

\textbf{Model Interpretability:}  
Developing interpretability tools, such as attention visualization and feature attribution methods, to enhance transparency and trust in operational forecasting systems.

\section*{Acknowledgment}
This research was conducted independently. The author gratefully acknowledges the Indian Climate and Energy Dashboard (ICED) and the NASA POWER project for providing the datasets used in this study. The author also thanks Prof. Sandeep Kaur, Department of Electrical Engineering, Punjab Engineering College (Deemed to be University), for academic guidance and constructive feedback during the early stages of this work. The author further thanks Lakshay Mittal, Peehu Sharma, Deepinder Singh, Abhipsit Bajpai, and Shivam Bansal for helpful discussions and assistance along the way.
\bibliographystyle{IEEEtran}
\bibliography{references}

\end{document}